\documentclass[runningheads]{llncs}

\usepackage{eccv}
\usepackage{eccvabbrv}
\usepackage{graphicx}
\usepackage{booktabs}
\usepackage{float}
\usepackage[protrusion=true,expansion=false]{microtype}
\usepackage{listings}
\usepackage{multirow}
\usepackage{xspace}
\newcommand{\ours}{DILLO\xspace}

\usepackage[breaklinks,colorlinks,citecolor=eccvblue]{hyperref}

\begin{document}
\title{Describe-Then-Act: Proactive Agent Steering via Distilled Language-Action World Models}
\titlerunning{Describe-Then-Act}

\author{
    Massimiliano Pappa\inst{1,2}\thanks{Equal contribution},
    Luca Romani\inst{1}\textsuperscript{*},
    Valentino Sacco\inst{1}\textsuperscript{*},\\ 
    Alessio Palma\inst{1},
    Stéphane Lathuilière\inst{2},
    Fabio Galasso\inst{1},\\ 
    Xavier Alameda-Pineda\inst{2},    
    Indro Spinelli\inst{1}
}

\authorrunning{M.~Pappa et al.}

\institute{
    Sapienza University of Rome, Italy \and
    Inria, Univ. Grenoble Alpes, CNRS, LJK
}

\maketitle

\begin{abstract}
Deploying safety-critical agents requires anticipating the consequences of actions before they are executed. While world models offer a paradigm for this proactive foresight, current approaches relying on visual simulation incur prohibitive latencies, often exceeding several seconds per step. In this work, we challenge the assumption that visual processing is necessary for failure prevention. We show that a trained policy's latent state, combined with its planned actions, already encodes sufficient information to anticipate action outcomes, making visual simulation redundant for failure prevention.
To this end, we introduce \textbf{DILLO} (\textbf{DI}sti\textbf{LL}ed Language-Acti\textbf{O}n World Model), a fast steering layer that shifts the paradigm from ``simulate-then-act'' to ``describe-then-act.'' DILLO is trained via cross-modal distillation, where a privileged Vision Language Model teacher annotates offline trajectories and a latent-conditioned Large Language Model student learns to predict semantic outcomes. This creates a text-only inference path, bypassing heavy visual generation entirely, achieving a 14$\times$ speedup over baselines. Experiments on MetaWorld and LIBERO demonstrate that DILLO produces high-fidelity descriptions of the next state and is able to steer the policy, improving episode success rate by up to 15\,pp and 9.3\,pp on average across tasks. Code is available at \href{https://github.com/MaxPappa/DILLO}{\texttt{github.com/MaxPappa/DILLO}}. 
\end{abstract}

\section{Introduction}
\label{sec:intro}

AI-driven agents are increasingly deployed in settings that demand high reliability, from robotic manipulation to autonomous navigation. However, a critical gap remains: these agents typically operate as ``black boxes'', selecting and executing actions without explicitly anticipating the consequences. In classical control theory, the standard solution is Model Predictive Control (MPC)~\cite{GARCIA1989335}: simulate the system forward and commit only to actions whose predicted outcomes are acceptable. The same principle underlies model-based reinforcement learning~\cite{sutton1991dyna, ha2018recurrent}, where world models forecast future states to steer policies.

\begin{figure}[!tb]
    \centering
    \includegraphics[width=0.98\linewidth]{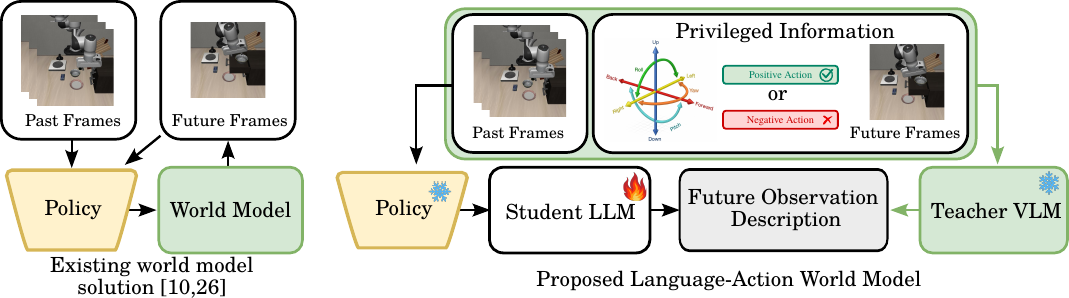}
\caption{(\textit{left}) Standard approaches rely on heavy visual world models to simulate future outcomes using expensive rendering. In contrast, (\textit{right}) DILLO avoids this visual dependency entirely. By distilling the foresight of a simulator-privileged Teacher VLM into a fast Language-Action World Model, DILLO predicts future outcome descriptions and success verdicts directly from policy latents ($z_t$) and action chunks ($a_{t:t+k}$). This shift from ``simulate-then-act'' to ``describe-then-act'' enables fast proactive steering on consumer hardware.}
    \label{fig:Figure 1}
\end{figure}

However, deploying world models in the tight control loops of dynamic agents presents a stark dilemma between foresight and latency. Current approaches generally fall into two categories. The first, post-hoc analysis \cite{duan2024aha, liu2023reflect, sinhareal}, operates as a reactive wrapper, diagnosing failures only after they occur. While computationally inexpensive, this approach is fundamentally incapable of prevention. The second paradigm, proactive visual simulation, utilizes heavyweight world models to generate high-dimensional future states (images/videos or their latent representation) \cite{hafner2023mastering, li2025roboticworldmodelneural}. While effective at anticipating risks, these ``visual'' world models incur prohibitive computational overhead. For instance, recent steering methods report inference latencies of nearly 4 seconds per decision \cite{wu2025foresight} on an enterprise RTX A6000 with 48GB of VRAM, rendering them impractical for real-time control on an embodied agent. We posit that for real-time near-future failure prevention, simulating the visual world is redundant. A policy's internal latent representation is explicitly trained to retain task-critical features: object geometry, relative distances, and contact dynamics. If this representation already contains the information needed to predict whether an action will succeed, then why pay the cost of re-rendering pixels? We call this the \textit{Latent Sufficiency Hypothesis}.

\noindent
To empirically verify this hypothesis, we propose \textbf{DILLO}(\textbf{DI}sti\textbf{LL}ed Language-Acti\textbf{O}n World Model), see Fig. \ref{fig:Figure 1}, a real-time reliability layer that shifts the paradigm from ``simulate-then-act'' to ``describe-then-act.'' DILLO is trained via cross-modal knowledge distillation: a privileged Vision-Language-Model (VLM) Teacher annotates offline trajectories with semantic outcomes, having access to the simulation environment. The Student (DILLO) then learns to predict these outcomes directly from the policy's compact latent state and candidate action chunk, without accessing raw visual observations. Architecturally, we leverage the reasoning capacity of Small Language Models, mapping projected latents directly to the language decoder. This vision-free design enables DILLO to perform fast inference, effectively decoding the agent's internal ``state of mind'' into human-readable foresight.

Specifically, DILLO generates two outputs: \textbf{(i) a natural language behavior preview} ($d$) that describes the expected physical interaction (\eg, \textit{``The gripper goes left and forward, starting to approach the ball''}), and \textbf{(ii) a binary verdict} ($c \in \{\textsc{Positive}, \textsc{Negative}\}$). A \textit{Negative} action is one causing stagnation, erratic movement, or progression toward failure; a \textit{Positive} action advances the task. This enables a dual-purpose mechanism: the verdict allows the agent to autonomously reject negative actions via rejection sampling, while the description provides interpretable information for human supervisors.

Our main contributions are:

\begin{itemize}
    \item We propose the \textit{Latent Sufficiency Hypothesis} and demonstrate empirically that a policy's latent state coupled with the sampled actions is a sufficient statistic for predicting failure-critical outcomes, making visual simulation redundant for proactive failure prevention.
    \item We introduce DILLO, a Distilled Language-Action World Model that distills a VLM teacher into a latent-conditioned LLM Student,
    shifting the paradigm from ``simulate-then-act'' to ``describe-then-act.'' This yields a ${\sim}14{\times}$ speedup over visual baselines~\cite{wu2025foresight}, enabling a full correction loop in \textbf{0.26\,s} on consumer hardware.
    \item \ours{}'s latent-based descriptions match or exceed the fidelity of vision-based baselines, including an oracle with access to future observations. Furthermore, we demonstrate effective steering on single- and multi-task policies across MetaWorld and LIBERO, improving episode success rate by up to \textbf{15\,pp} and \textbf{9.3\,pp} on average across tasks, achieving a verdict classification accuracy of 91.4\% without access to any visual observation at inference time.

\end{itemize}

\section{Related Work}\label{sec:related_works}

We propose a real-time, language-based failure detection layer within the context of three dominant paradigms: reactive failure analysis, proactive world modeling, and action-language grounding.

\noindent\textbf{Reactive Post-Hoc Analysis.}
The most common approach to agent interpretability follows an ``act-then-describe'' paradigm, analyzing failures only \textit{after} they have occurred. Methods such as Aha~\cite{duan2024aha} and REFLECT~\cite{liu2023reflect} leverage Vision-Language Models (VLMs) to process offline trajectory data, generating human-readable summaries of past errors. Other works treat VLMs as behavioral critics~\cite{guan2024task} or as a means to ground historical robotic experiences within a linguistic framework~\cite{wangcan}. While runtime monitors~\cite{agiaunpacking, sinhareal} operate closer to the point of execution, they are primarily designed to detect anomalies or consistency violations post-occurrence. Similarly, SAFE~\cite{gu2025safe} analyzes the latent states of Vision-Language-Action models (VLAs) to predict task success or failure in a task-agnostic manner. Although these methods are indispensable for data curation and debugging, they remain fundamentally reactive, offering a ``failure autopsy'' rather than a preemptive intervention. In physical environments where robotic actions are often irreversible, such retrospective detection cannot substitute for robust, real-time prevention.

\noindent\textbf{Proactive World Models for Control.}
To prevent failures, an agent must anticipate them. The concept of \textit{World Models} champions this proactive paradigm, evolving from early visual foresight~\cite{finn2016unsupervised, finn2017deep, ebert2018visual} to powerful latent dynamics models like Dreamer~\cite{hafner2019dream, hafner2020mastering, hafner2023mastering}. While these models effectively hallucinate future rewards for policy learning, they typically output latent vectors that are opaque to human supervisors. 
Recently, this paradigm was extended to \textit{semantic} policy steering by Forewarn~\cite{wu2025foresight}, which utilizes VLMs to narrate and evaluate simulated latent states. While validating the utility of linguistic previews, such ``VLM-in-the-loop'' methods remain computationally prohibitive. Relying on visual encoders to process predicted latent vectors and their subsequent evaluation incurs high latency (\eg, $\sim$3.7 seconds per decision~\cite{wu2025foresight}), rendering them impractical for the tight control loops of dynamic agents.
\ours{} takes a different approach: rather than performing expensive, future-state simulation, we distill semantic foresight into a fast, latent-conditioned predictor enabling real-time safety without the associated simulation and evaluation overhead.

\noindent\textbf{Action-Language Grounding.}
Finally, we distinguish our work from rationale-driven Explainable AI (XAI)~\cite{milani2024explainable, wojciechowski2024faithful, sammani2022nlx}. Methods like Embodied Chain-of-Thought (ECoT)~\cite{zawalski24ecot} use language to expose an agent's \textit{internal logic} (\eg, ``I need the cup, so I will move my arm''). In contrast, \ours{} is outcome-driven; it does not explain the agent's reasoning but rather predicts the external physical consequence of an action (\eg, ``This action will cause the gripper to move backward, increasing its distance from the cup''). This distinction allows \ours{} to function as a controller-agnostic safety module that can be layered onto any policy to provide a fast, interpretable sanity check.

\section{Distilled Language Action World Model}
\label{sec:approach}

\begin{figure}[!tb]
    \centering
    \includegraphics[width=0.99\linewidth]{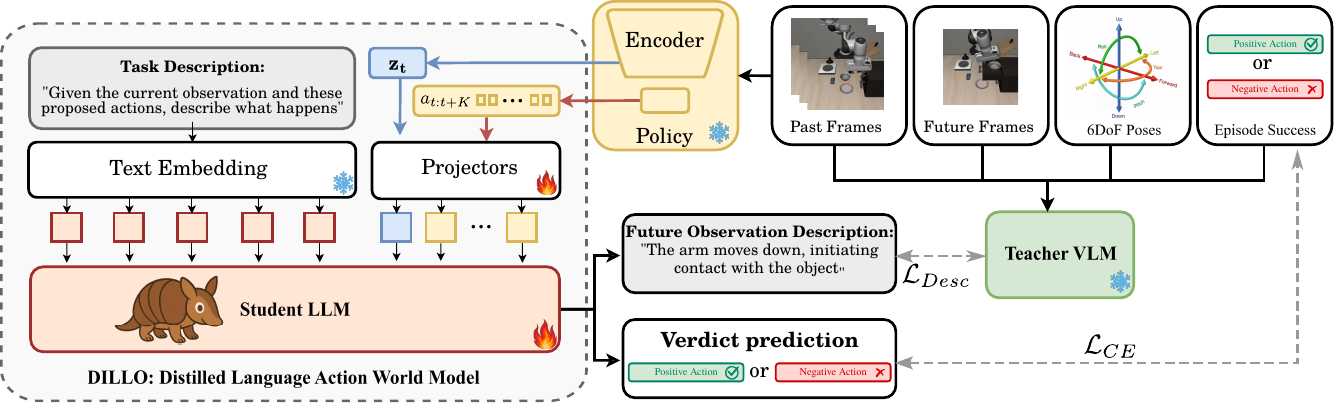}
    \caption{\textbf{DILLO Training Pipeline.} We distill the foresight of a VLM teacher. The pipeline bypasses visual processing by projecting the fixed policy's internal latents ($z_t$) and action chunks ($a_{t:t+k}$) directly into the LLM's embedding space.}
    \label{fig:Figure 2}
\end{figure}

We introduce \textbf{DILLO} (\textbf{DI}sti\textbf{LL}ed Language-Acti\textbf{O}n World Model), a framework for real-time failure prevention that decouples semantic anticipation from visual simulation. We formulate the problem as cross-modal knowledge distillation, where a lightweight student learns to approximate the semantic foresight of a privileged teacher solely from the policy's internal representations and predicted actions, see Fig. \ref{fig:Figure 2}.

\subsection{Semantic Anticipation}
Consider a robotic agent operating in a Partially Observable Markov Decision Process (POMDP). At decision step $t$, the agent receives an observation $o_t \in \mathcal{O}$ and processes it via the policy's encoder $\mathcal{E}_{\pi}$ to extract a compact latent state:
\begin{equation}
    z_t = \mathcal{E}_{\pi}(o_t) \in \mathbb{R}^{D_z}
\end{equation}
Conditioned on this latent state, the policy proposes a sequence of $k$ future actions, the action chunk~\cite{liu2025bidirectional} $a_{t:t+k}$.
Our objective is to learn a \textbf{Language-Action World Model} modeling a distribution $P_{\theta}(y_{t+k} \mid z_t, a_{t:t+k})$ that predicts the semantic outcome $y_{t+k}$ directly from these internal representations, without accessing the raw visual observation. Standard world models use the state representation, often an RGB image, to predict high-dimensional future observations $\hat{o}_{t+k}$ (pixel space).

The semantic outcome $y_{t+k}$, consists of a natural language description $d_{t+k}$ (\eg, physical dynamics, contacts) and a verdict $c_{t+k} \in \{\textsc{Positive}, \textsc{Negative}\}$ indicating whether the action chunk advances or hinders the task. The distribution of this outcome is conditioned on the environment dynamics:
\begin{equation}
    P_\mathit{env}(y_{t+k} \mid o_t, a_{t:t+k})
\end{equation}
This is estimated by a teacher distribution $P_{\mathcal{T}}(y_{t+k} \mid o_{t:t+k})$ that has privileged access to a world model or simulator:
\begin{equation}
    P_{\mathit{env}}(o_{t+k} \mid o_{t:t+k-1}, a_{t:t+k})
\end{equation}
The teacher receives the full ground-truth visual history from the simulated environment, along with 6DoF poses for both the object and the gripper.

\subsection{The Latent Sufficiency Hypothesis}
Evaluating $P_{\mathcal{T}}$ is computationally prohibitive for real-time control. Our objective is to learn a fast estimator, the student distribution $P_{\theta}(y_{t+k} \mid z_t, a_{t:t+k})$, conditioned only on the policy's latent state and planned actions. The feasibility of this approach rests on the following hypothesis:

\noindent \textbf{Hypothesis on Latent Sufficiency.} \textit{For the task of semantic world modeling, the mutual information between the policy latent-action tuple $(z_t,\, a_{t:t+k})$ and the future outcome $y_{t+k}$ approximates the mutual information between the full observation history $o_{t:t+k}$ and $y_{t+k}$:}

\begin{equation}
    I\bigl (o_{t:t+k};\, y_{t+k}\bigr ) \;\approx\; I\bigl ((z_t,\, a_{t:t+k});\, y_{t+k}\bigr ). 
    \label{eq}
\end{equation}

Whether trained via Reinforcement Learning (\eg, SAC \cite{Haarnoja2018SoftAO}) to maximize task reward, or Imitation Learning (\eg, ACT \cite{zhao2023act}) to reconstruct expert action intent, the encoder $\mathcal{E}_{\pi}$ must extract and retain interaction-critical representations, such as object geometry, relative distances, and contact dynamics. We claim that $(z_t,\, a_{t:t+k})$, derived solely from $o_t$, carries as much predictive information about $y_{t+k}$ as the full observation history seen by the teacher, making $z_t$ a sufficient statistic for $y_{t+k}$ and rendering generating and re-encoding raw pixels unnecessary. This sufficiency extends to failing policies. When a policy encounters an out-of-distribution state, the resulting latent $z_t$ captures the ``context of failure'' (\eg, uncertainty or feature mismatch), and DILLO learns to map these latents to Negative verdicts. 

By minimizing the Kullback--Leibler divergence between the teacher and student distributions ~\cite{hinton2015distillingknowledgeneuralnetwork}:

\begin{equation}
    \min_{\theta}\; D_{KL}\!\left( P_{\mathcal{T}}(y_{t+k} \mid o_{t:t+k}) \;\|\; P_{\theta}(y_{t+k} \mid z_t, a_{t:t+k}) \right)
\end{equation}
we distill the teacher's future-aware foresight into a Language-Action World Model that translates the policy's existing manifold into natural language descriptions. This reduces to a cross-entropy minimization over $(z_t, a_{t:t+k})$. DILLO thereby predicts outcomes over the horizon $t \to t+k$ with negligible computational overhead.

\subsection{Models}\label{sec:models}

\noindent\textbf{Teacher.} We instantiate the teacher $\mathcal{T}$ as a VLM with access to privileged information from the simulator state, eliminating the depth ambiguity and occlusion issues inherent to raw vision. In particular, the model has access to:

\begin{enumerate}
    \item \textbf{Visual Context:} The sequence of RGB observations corresponding to the execution of the action chunk.
    \item \textbf{Geometric Deltas:} Directional changes and 6DoF poses of the end-effector and task-relevant objects, grounding the VLM in the physics of the scene and mitigating the spatial hallucinations common in vision-only models.
    \item \textbf{Episode Outcome Signal:} A binary indicator of whether the specific action chunk led to successful task completion. 
\end{enumerate}

In physical deployment, this privileged information can be substituted by state-of-the-art 6D pose estimation algorithms~\cite{foundationposewen2024}. 

To construct the distillation dataset, the privileged teacher model processes the ground-truth state information alongside the natural language task description and the target goal configuration. We segment policy rollouts at the state-action level. For each transition, the teacher generates a natural language description $d_{\mathcal{T}}$ and assigns a binary verdict $c_{\mathcal{T}} \in \{\textsc{Positive}, \textsc{Negative}\}$. A \textsc{Negative} verdict explicitly penalizes trajectories that exhibit stagnation, erratic actuation, or behaviors otherwise detrimental to task progression. Finally, we pair these high-level semantic labels with the base policy's internal latent state $z_t$ and the predicted action chunk $a_{t:t+k}$. This yields the final distillation tuple:

$$\tau = (z_t, a_{t:t+k}, d_{\mathcal{T}}, c_{\mathcal{T}})$$

\noindent\textbf{Student (DILLO).}
The student $f_{\theta}$ parameterizes $P_{\theta}$ using the Gemma \cite{gemmateam2025gemma3technicalreport} family of open weights. We implement two variants to validate scalability:
\begin{itemize}
    \item \textbf{DILLO-1B:} Uses the standard language-only Gemma-1B-it LLM backbone.
    \item \textbf{DILLO-4B:} 
    Unlike standard VLM inference, which relies on a 417M-parameter SigLIP encoder, our model uses only the large language model of the Gemma-VLM-4B-it architecture. This ensures that DILLO-4B benefits from the reasoning capacity of a VLM while incurring zero visual overhead.
\end{itemize}

To map the continuous control space to the language space, we employ two learnable projectors:
\begin{itemize}
    \item \textbf{Latent Projector} $P_z: \mathbb{R}^{D_z} \to \mathbb{R}^{D_{emb}}$, mapping the frozen policy embedding to the LLM input space.
    \item \textbf{Action Projector} $P_a: \mathbb{R}^{D_a} \to \mathbb{R}^{D_{emb}}$, mapping each action chunk $a_{t:t+k}$ to an action vector that is then fed to the LLM.
\end{itemize}
The input sequence to the Transformer is constructed as:

\begin{equation}
    X_{in} = \texttt{[TASK PROMPT]} \oplus P_z(z_t) \oplus P_a(a_{t:t+k})
\end{equation}

Here, $X_{in}$ contains no tokens derived from $o_t$, ensuring independence from the visual encoder during inference.

\subsection{Training}
\label{sec:multi_obj_opt}

End-to-end mapping of continuous control latents into discrete LLM tokens is prone to modality collapse. To stably ground the LLM in the policy's physical manifold, we employ a progressive three-stage curriculum:

\textbf{Stage 1: Projector Alignment.} To bridge the dimensional gap without destabilizing pre-trained weights, we freeze the LLM backbone $\theta_{LLM}$ and optimize only the linear projectors $\{P_z, P_a\}$ \cite{liu2023llava}. We use standard autoregressive next-token prediction on the teacher's descriptions to ensure the projected latents act as valid prompt embeddings.

\textbf{Stage 2: Description Distillation.} To establish a dense semantic prior, we apply Low-Rank Adaptation (LoRA) \cite{hu2022lora} to the LLM. We jointly optimize the LoRA parameters and projectors to reconstruct the teacher's natural language rationale $d_{\mathcal T}$:
\begin{equation}
\mathcal{L}_{Desc} = -\mathbb{E}_{\mathcal{D}} \left[ \sum_{i=1}^{|d|} \log p_{\theta}(d_i | d_{<i}, z_t, a_{t:t+k}) \right]
\end{equation}

\textbf{Stage 3: Verdict Optimization.} Finally, we introduce the binary task constraint. The model generates the rationale followed by a verdict token $\hat{c} \in \{\textsc{Positive}, \textsc{Negative}\}$. We minimize the joint objective: 
\begin{equation}
    \mathcal{L}_{Total} = \mathcal{L}_{Desc} + \lambda \cdot \mathcal{L}_{CE}(c_{\mathcal{T}}, \hat{c})
\end{equation}
where $\mathcal{L}_{CE}$ is the cross-entropy against the ground-truth verdict $c_{\mathcal{T}}$, ensuring the natural language descriptions are grounded with the task execution.

\subsection{Proactive Steering via Latent Rejection Sampling}\label{sec:steering}

\begin{figure}[!tb]
  \centering
  \includegraphics[width=0.9\textwidth]{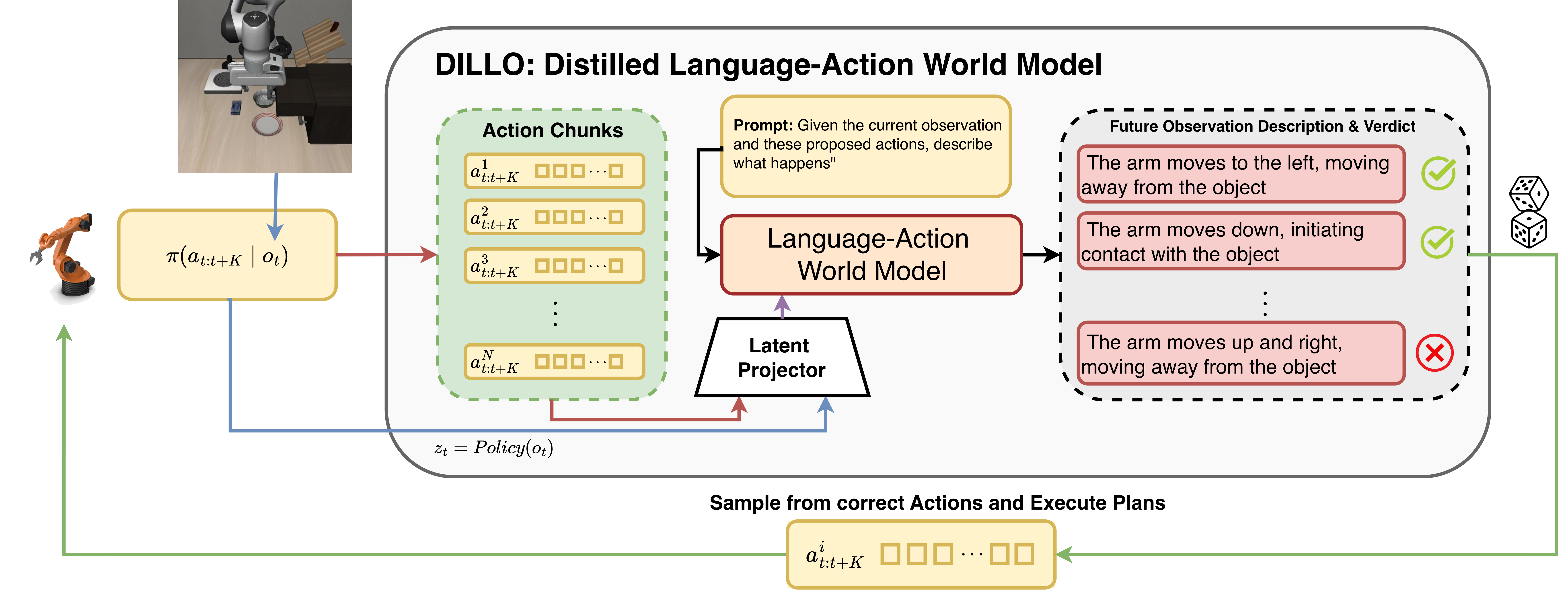}
  \caption{\textbf{Proactive Policy Steering via Latent Rejection Sampling.} At inference time, \ours{} acts as a high-speed safety filter. The policy proposes $N$ candidate action chunks ($a^i_{t:t+k}$). Instead of re-rendering future images for each candidate, \ours{} projects the policy latent ($z_t$) directly into the Language-Action World Model. The model predicts a semantic description and verdict for each chunk. Negative trajectories (red $\times$) are rejected, and the first verified positive plan (green $\checkmark$) is executed.}
  \label{Figure 3}
\end{figure}

DILLO's distilled verdict $\hat{c}$ enables it to function as a controller-agnostic, zero-overhead safety filter at inference time. Rather than executing the first proposed action chunk, the agent uses DILLO to screen a batch of $N$ candidate plans before committing to any physical action, as illustrated in Fig.~\ref{Figure 3}.

\noindent \textbf{Inference Protocol.} At each decision step $t$, the base policy encodes the current observation into a shared latent state $z_t = \mathcal{E}_{\pi}(o_t)$, computed \emph{once} regardless of how many candidates are subsequently evaluated. The policy then proposes a budget of $N$ candidate action chunks by sampling independently from its action distribution:
\begin{equation}
    \bigl\{ a^i_{t:t+k} \bigr\}_{i=1}^{N} \sim \pi(\cdot \mid z_t)
\end{equation}
DILLO evaluates all $N$ candidates in a single batched forward pass. For each candidate, the shared latent $z_t$ and an encoded projection of the proposed action sequence are jointly fed into the model:
\begin{equation}
    y^i_{t+k} = \bigl(d_{t+k}^i,\, c_{t+k}^i\bigr) = f_\theta\!\left(P_z(z_t),\, 
    P_a(a^i_{t:t+k})\right), \quad i = 1, ... N
\end{equation}
The agent then selects any candidate that receives a \textsc{Positive} verdict. If no \textsc{Positive} verdict is found within the budget, the agent falls back to executing a candidate from the initial proposal, ensuring the control loop is never stalled. In our experiments we set $N{=}5$. Notably, the framework naturally supports multiple resampling rounds: the policy can draw and evaluate successive batches of $N$ candidates up to $R$ rounds (where $R$ is a configurable budget), allowing up to $R \cdot N$ total evaluations before fallback at the cost of added latency (see Sec.~\ref{subsec:inference_latency}).

\section{Experiments}
\label{sec:experiments}

Our evaluation validates two central claims. First, \textit{latent sufficiency}: 
does DILLO, conditioned solely on latent states ($z_t$), produce semantic 
descriptions ($y_{t+k}$) that faithfully match ground-truth physical outcomes 
($o_{t+k}$)? Second, \textit{utility}: does DILLO's verdict effectively steer 
a base policy toward higher episode success rates? We conduct experiments in 
\textbf{MetaWorld}~\cite{yu2020meta} (\texttt{Soccer}, \texttt{Sweep-Into}, 
\texttt{Drawer-Open}) and \textbf{LIBERO}~\cite{liu2023libero} 
(\texttt{Goal}, \texttt{Object}, \texttt{Spatial}, \texttt{10}, \texttt{90}).

\subsection{Constructing the Distillation Dataset}
\label{sec:dataset}

To train DILLO, we collect distillation datasets by rolling out pre-trained policies across two distinct algorithmic paradigms, capturing the latent transitions inherent to both specialized and generalist behaviors.

\noindent \textbf{Single-Task Reinforcement Learning (MetaWorld).} We roll out SAC-trained policies~\cite{Haarnoja2018SoftAO} across three manipulation tasks (Soccer, Sweep-Into, Drawer-Open). Each policy is specialized for a single 
task; correspondingly, a separate DILLO model is distilled per task. The collected trajectories consist of purely observational latents, $z_t = \mathcal{E}(o_t)$, where the representation is driven entirely by the RL reward signal, with no language conditioning.

\noindent \textbf{Multi-Task Imitation Learning (LIBERO).} We roll out action-chunking, imitation-learned policies \cite{zhao2023act} across five task suites (LIBERO-Goal, LIBERO-Object, LIBERO-Spatial, LIBERO-10, LIBERO-90). Here, a single policy must generalize over a shared set of language conditioned tasks, and the distillation dataset reflects this diversity. A single DILLO model is therefore distilled from trajectories spanning heterogeneous, language-specified goals, testing whether the latent space of a generalist policy retains sufficient structure for cross-task outcome prediction.

\noindent \textbf{Failures.} A reliable predictive model must identify failure modes across all levels of agent capability. A classifier trained solely on expert successes and random failures will fail to generalize to subtle execution errors. To guarantee a comprehensive distribution of both catastrophic and marginal failures, we curate our dataset using two complementary strategies:

\begin{itemize}
    \item \textbf{Spectrum of Competency Sampling.} Rather than relying exclusively on converged experts, we sample rollouts from policy checkpoints saved at distinct performance thresholds. We incorporate low-competency models ($20\%$ success) to capture erratic exploratory behaviors, mid-competency models ($50\%$ success) to provide nuanced ``near-miss'' examples \cite{agiaunpacking} that are difficult to classify, and high-competency models ($80\%$ success) to establish ground-truth task alignment while offering ``hard negatives'' through their rare failures.

    \item \textbf{Perception Noise Injection.} During the data collection step, we inject Gaussian noise $\epsilon \sim \mathcal{N}(0, \sigma^2)$ into the observation $o_t$ before querying the policy $\pi(a_{t:t+k} | o_t + \epsilon)$. This represents a standard proxy for sensor uncertainty used in Sim-to-Real domain randomization. The perturbation causes the agent to misinterpret its state, inducing trajectory drift and compounding execution errors. Crucially, this generates failures that originate from a competent action manifold rather than random initialization, systematically mimicking the covariate shift commonly observed during physical deployment.
\end{itemize}

\subsection{Metrics}
\label{sec:fidelity_metrics}
To support the hypothesis on ``latent sufficiency'', we must quantify the alignment between the text predicted from latents ($z_t$) and the actual physical outcome observed in the simulator ($o_{t+k}$). 
We compare \ours{} against baselines backed by Gemma-3-1B and Gemma-3-4B:
\begin{itemize}
    \item \textbf{Zero-Shot (ZS):} The model receives the current observation $o_t$ and the task description.
    \item \textbf{Few-Shot (FS):} The model receives $o_t$ along with $3$ in-context examples of ($o_t$, $y_{t+k}$) pairs.
    \item \textbf{Few-Shot Captioning (CAP):} The model receives the current observation $o_t$, the \textit{future} observation $o_{t+k}$, and $3$ in-context examples of ($o_t$,$o_{t+k}$, $y_{t+k}$) tuples. It does not need to predict the future; it can caption what actually happened therefore it should be seen as an oracle.
\end{itemize}

To evaluate baselines and \ours, we rely on two  fidelity metrics:

\noindent\textbf{Text-to-Observation Fidelity (T2O).} To quantify low-level dynamic accuracy without the noise of linguistic ambiguity, we map both physical state transitions and generated descriptions into a shared space of canonical directional labels. A deterministic function, $f_{extract}(o_t, o_{t+k})$, computes the true displacement for the gripper and objects, as well as relative relational changes (\eg, \textit{approaching} vs. \textit{receding}) across three orthogonal axes. Axes with negligible movement are assigned a ``static'' label. Simultaneously, a parser $g_{parse}(d)$ projects the natural language prediction into the same canonical set via robust synonym mapping. The final T2O score is computed as the accuracy of matched labels across all dynamic variables, serving as a direct proxy for physical grounding (see supplementary material for the definition of $g_{\text{parse}}$).

\noindent\textbf{Reasoning-Based Semantic Score.}
While T2O captures dynamics, it may miss nuanced semantic details. We employ a Judge LLM (Qwen2.5-32B-Instruct \cite{qwen2025qwen25technicalreport}) to score the semantic alignment between \ours{}'s prediction $d$ and the ground-truth reference $d_{\mathcal{T}}$.
The judge decomposes the reference into a set of atomic facts $F = \{f_i\}_{i=1}^{|F|}$. Subsequently, it evaluates the extent to which the candidate description $d$ entails each fact, assigning a support score $m_i \in \{0, 0.5, 1\}$ (contradiction, partial implication, and full support). We define the final metric as the mean fact recall:
\begin{equation}
    \text{Score}(d_{t+k} | d_{\mathcal{T}}) = \frac{1}{|F|} \sum_{i=1}^{|F|} m_i
\end{equation}
This metric offers a robust, interpretable measure of semantic coverage beyond simple lexical overlap metrics like BLEU~\cite{papineni2002bleu} or ROUGE~\cite{lin2004rouge}.

\noindent\textbf{Episode Success Rate (ES).}
We measure the percentage of episodes where the agent completes the task goal. We report the success rate of the \ours{}-Steered Policy versus the Base Policy (no steering). A high ES for \ours{} confirms that the distilled verdict $\hat{c}$ is a reliable signal for filtering negative actions.

\begin{table}[t]
    \centering
    \caption{\textbf{Fidelity Results.} We measure the alignment between the predicted description and the actual simulator state. We compare \ours{} against LLM and VLM baselines (Few-Shot, Zero-Shot, and CAP) across two model scales (1B and 4B parameters) on both the MetaWorld and LIBERO suites. \ours{} consistently achieves high fidelity and outperforms or matches baselines, validating that the latent state is a sufficient statistic for semantic prediction.}
    \label{tab:fidelity_results_merged}
    \setlength{\tabcolsep}{4.5pt}
    \resizebox{\textwidth}{!}{
    \begin{tabular}{llcccccccc}
        \toprule
        & & \multicolumn{4}{c}{\textbf{1B Parameter Models}} & \multicolumn{4}{c}{\textbf{4B Parameter Models}} \\
        \cmidrule(lr){3-6} \cmidrule(lr){7-10}
        \textbf{Environment} & \textbf{Task} & \textbf{\ours-1B} & FS & ZS & CAP & \textbf{\ours-4B} & FS & ZS & CAP \\
        \midrule
        \multirow{3}{*}{MetaWorld} 
        & Soccer      & \textbf{0.518} & 0.495 & 0.398 & 0.477 & 0.586 & 0.501 & 0.540 & \textbf{0.626} \\
        & Sweep-Into  & \textbf{0.562} & 0.329 & 0.528 & 0.271 & 0.554 & 0.630 & 0.687 & \textbf{0.725} \\
        & Drawer-Open & \textbf{0.706} & 0.626 & 0.419 & 0.612 & \textbf{0.705} & 0.555 & 0.553 & 0.609 \\
        \midrule
        \multirow{5}{*}{LIBERO} 
        & Goal    & \textbf{0.727} & 0.262 & 0.287 & 0.723 & 0.710 & 0.422 & 0.241 & \textbf{0.905} \\
        & Object  & 0.622 & 0.388 & 0.204 & \textbf{0.820} & 0.587 & 0.546 & 0.225 & \textbf{0.877} \\
        & Spatial & 0.668 & 0.546 & 0.341 & \textbf{0.794} & 0.656 & 0.563 & 0.194 & \textbf{0.902} \\
        & 10      & \textbf{0.648} & 0.214 & 0.199 & 0.630 & 0.686 & 0.449 & 0.265 & \textbf{0.867} \\
        & 90      & \textbf{0.708} & 0.258 & 0.206 & 0.640 & 0.650 & 0.278 & 0.258 & \textbf{0.713} \\
        \bottomrule
    \end{tabular}
    }
\end{table}

\begin{table}[t]
    \centering
    \caption{\textbf{Reasoning Based LLM Score.} We compare \ours{} against VLM baselines (Few-Shot, Zero-Shot, and CAP) across two model scales (1B and 4B parameters). \ours{} consistently outperforms standard baselines in semantic reasoning, even approaching or exceeding the CAP in the 1B regime.}
    \label{tab:llm_results_merged}
    \setlength{\tabcolsep}{5pt}
    \resizebox{\textwidth}{!}{
    \begin{tabular}{llcccccccc}
        \toprule
        & & \multicolumn{4}{c}{\textbf{1B Parameter Models}} & \multicolumn{4}{c}{\textbf{4B Parameter Models}} \\
        \cmidrule(lr){3-6} \cmidrule(lr){7-10}
        \textbf{Environment} & \textbf{Task} & \textbf{\ours-1B} & FS & ZS & CAP & \textbf{\ours-4B} & FS & ZS & CAP \\
        \midrule
        \multirow{3}{*}{MetaWorld} 
        & Soccer      & \textbf{0.471} & 0.301 & 0.204 & 0.277 & \textbf{0.504} & 0.361 & 0.328 & 0.425 \\
        & Sweep-Into  & \textbf{0.592} & 0.372 & 0.284 & 0.388 & \textbf{0.607} & 0.391 & 0.381 & 0.364 \\
        & Drawer-Open & \textbf{0.655} & 0.510 & 0.189 & 0.427 & \textbf{0.628} & 0.497 & 0.212 & 0.252 \\
        \midrule
        \multirow{5}{*}{LIBERO} 
        & Goal    & \textbf{0.834} & 0.259 & 0.271 & 0.686 & \textbf{0.830} & 0.487 & 0.278 & 0.816 \\
        & Object  & \textbf{0.803} & 0.380 & 0.208 & 0.668 & \textbf{0.723} & 0.519 & 0.243 & 0.684 \\
        & Spatial & \textbf{0.782} & 0.448 & 0.323 & 0.684 & 0.758 & 0.545 & 0.282 & \textbf{0.792} \\
        & 10      & \textbf{0.822} & 0.227 & 0.205 & 0.611 & \textbf{0.797} & 0.493 & 0.282 & 0.777 \\
        & 90      & \textbf{0.708} & 0.258 & 0.206 & 0.640 & 0.650 & 0.278 & 0.258 & \textbf{0.713}\\
        \bottomrule
    \end{tabular}
    }
\end{table}

\subsection{Validating Latent Sufficiency}
\label{subsec:fidelity}

\begin{figure}[!htbp]
    \centering
    \includegraphics[width=0.85\linewidth,height=7cm,keepaspectratio]{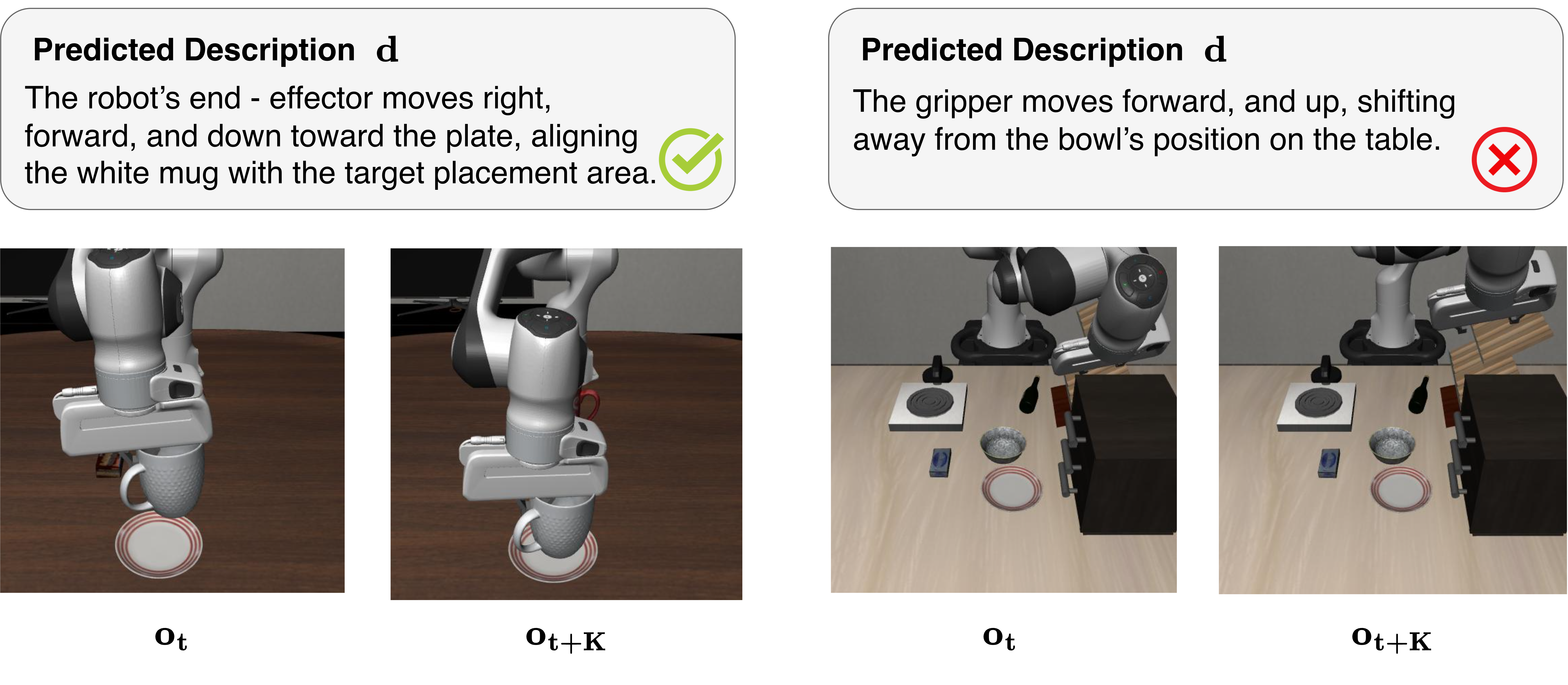}
    \caption{\textbf{Qualitative validation of \ours{}'s foresight for `Positive' (left) and `Negative' (right) outcomes.} \ours{} generates the predicted description ($d$) and verdict (check/cross) using only the policy's latent state ($z_t$) and the proposed action chunk. The initial observation ($o_t$) and ground-truth future ($o_{t+K}$) are shown purely for visual comparison, confirming the high fidelity of the latent-based prediction.}
    \label{fig:qualitative}
\end{figure}

We validate the Latent Sufficiency Hypothesis using the T2O fidelity and LLM-judged semantic scores reported in Tables~\ref{tab:fidelity_results_merged} and~\ref{tab:llm_results_merged}. The results reveal three consistent trends across both benchmarks.

\noindent\textbf{Latent beats pixels.}
The ``blind'' \ours{}-4B model consistently and significantly outperforms the pixel-based Few-Shot and Zero-Shot baselines on both MetaWorld and LIBERO, confirming that the policy latent $z_t$ is a more informative predictor of future dynamics than the current visual observation $o_t$.

\noindent\textbf{Latent beats the future image.}
Remarkably in Table \ref{tab:fidelity_results_merged}, \ours{}-4B's T2O score also exceeds the CAP baseline on \texttt{Drawer-Open} (\textbf{0.705} vs.\ \textbf{0.609}), where CAP has access to the actual future observation $o_{t+k}$ and $3$ in-context examples of ($o_t$,$o_{t+k}$, $y_{t+k}$), which allow for pure captioning without having to predict the future. This indicates that $z_t$ is not merely sufficient but a \emph{superior} predictive statistic: as a task-optimized embedding, it is less susceptible to the noise and perceptual ambiguity inherent in raw pixel data. This directly justifies the Zero-Visual-Overhead architecture.

\noindent\textbf{Semantically richer descriptions.}
The LLM-judged scores in Table~\ref{tab:fidelity_results_merged} show that both \ours{}-1B and \ours{}-4B outperform all baselines across the vast majority of tasks on both benchmarks. On MetaWorld \texttt{Drawer-Open}, \ours{}-4B achieves \textbf{0.628} against \textbf{0.497} for FS and \textbf{0.252} for CAP. This strong performance extends to LIBERO, with two notable exceptions. On LIBERO-90, which spans 90 distinct tasks, performance relative to CAP suggests that larger-scale data collection is needed to handle the increased task diversity. On LIBERO-Spatial, significant variability in object placements, combined with the absence of explicit goal position conditioning in the current \ours{} formulation, limits generalization to spatially complex configurations. These are targeted failure modes rather than systemic weaknesses, and both suggest clear directions for future work.

Figure~\ref{fig:qualitative} provides qualitative validation on LIBERO. In the \textsc{Positive} outcome (left), the predicted description $\hat{d}$ accurately articulates the arm's successful engagement with the object. In the \textsc{Negative} outcome (right), \ours{} correctly anticipates the impending failure, the arm disengages from the target, and issues a \textsc{Negative} verdict. Both the description and the verdict are generated solely from $z_t$ and $a_{t:t+k}$, with no access to visual observations at inference time.

\subsection{Proactive Policy Steering and Inference Latency}

\begin{table}[t]
    \centering
    \caption{\textbf{Steering Success Rates and Verdict Accuracy.} We report Steering Success Rates (SR) and Safety Verdict Accuracy (\%). High success rates and accuracy confirm that the model effectively steers rollouts while discriminating successful dynamics from failures.}
    \label{tab:combined_results}
    \footnotesize
    \setlength{\tabcolsep}{2.4pt}
    \begin{tabular}{ll ccc | cc}
        \toprule
        & & \multicolumn{3}{c}{\textbf{Success Rate}} & \multicolumn{2}{|c}{\textbf{Verdict Accuracy}} \\
        \cmidrule(lr){3-5} \cmidrule(lr){6-7}
        \textbf{} & \textbf{Task} & Base & \ours-1B & \ours-4B & \ours-1B & \ours-4B \\
        \midrule
        \multirow{3}{*}{MetaWorld} 
        & Soccer      & 65.0\% & 75.0\% & \textbf{80.0\%} & 82.7\% & \textbf{87.4\%} \\
        & Sweep-Into  & 80.0\% & \textbf{90.0\%} & 85.0\% & \textbf{97.7\%} & 92.1\% \\
        & Drawer-Open & 75.0\% & 70.0\% & \textbf{90.0\%} & 80.6\% & \textbf{94.7\%} \\
        \midrule
        \multirow{5}{*}{LIBERO} 
        & Goal    & 86.0\% & 87.5\% & \textbf{89.0\%} & 80.0\% & \textbf{87.5\%} \\
        & Object  & 82.0\% & 86.5\% & \textbf{88.0\%} & 79.8\% & \textbf{81.2\%} \\
        & Spatial & 70.0\% & \textbf{76.0\%} & 73.0\% & 77.8\% & \textbf{86.7\%} \\
        & 10      & 64.0\% & 71.5\% & \textbf{74.0\%} & 88.6\% & \textbf{92.5\%} \\
        & 90      & \textbf{71.4\%} & 68.6\% & 70.5\% & \textbf{86.9\%} & 85.7\% \\ 
        \bottomrule
    \end{tabular}
\end{table}
\noindent\textbf{Proactive Policy Steering.}
We employ \ours{} as a rejection sampling controller ($N{=}5$) to quantify steering gains over 20 episodes per task. Table~\ref{tab:combined_results} reports success rates across both benchmarks.

On MetaWorld, \ours{} consistently improves over the base policy. On  \texttt{Soccer}, \ours{}-4B raises the success rate from 65.0\% to \textbf{80.0\%}, and even the smaller \ours{}-1B achieves 75.0\%. On \texttt{Sweep-Into}, \ours{}-1B reaches \textbf{90.0\%}, the highest score across all model variants on that task.

On LIBERO, gains are consistent across the four main suites. \ours{}-4B improves LIBERO-Goal from 86.0\% to \textbf{89.0\%} and LIBERO-Object from 82.0\% to \textbf{88.0\%}, confirming that the distilled verdict generalizes effectively to multi-task, language-conditioned settings. The exception is LIBERO-90, where the base policy (71.4\%) marginally outperforms both DILLO variants (68.6\% and 70.5\%). We analyze 900 LIBERO-90 episodes under varying rejection thresholds to isolate the source of failures. Moderate gating yields a net positive effect, improving success by $+1.7\%$ overall and by $+17.5\%$  on tasks below $70\%$ baseline success. Heavier gating, however, over-rejects candidate actions and reduces success by $-4.4\%$, indicating that broad multi-task deployment requires task-conditioned calibration rather than a single global rejection threshold.

\noindent\textbf{Generalization across LIBERO suites.}
To test cross-suite transfer, we steer the LIBERO-10 ACT policy with our LIBERO-90-trained explainer. \ours{} raises policy success from 64.0\% to 69.5\%, approaching the 71.5\% of same-suite steering. Meaning that the learned semantic evaluator does not simply memorize a single suite but can generalize to similar scenarios.

\noindent\textbf{Positive/Negative Classification.} 
To understand the source of this effectiveness, we analyze the binary classification accuracy of the verdict token $\hat{c}$. Table~\ref{tab:combined_results} shows that \ours{}-4B achieves an average accuracy of \textbf{91.4\%} across tasks, without access to any visual observation at inference time.
A human validation study (see supplementary material) further corroborates these results. On MetaWorld, \ours{} agrees with human positive assessments on 84.1\% of 60 episodes. On LIBERO, \ours{}'s verdict agrees with ground truth at 72.1\%, exceeding the human-vs-ground-truth agreement of 69.5\%, while human evaluators rate its descriptions at $3.62/5.0$, confirming interpretability to non-expert observers.

\noindent\textbf{Inference Latency.}\label{subsec:inference_latency}
Per decision step (one action chunk of 20 simulation steps), the base policy accounts for $T_{\text{sample}} \approx 10$\,ms, and \ours{} adds 0.26\,s (1B) and 0.38\,s (4B) per step. This is a $\mathbf{{\sim}14\times}$ and $\mathbf{{\sim}10\times}$ speedup over Forewarn~\cite{wu2025foresight} (3.70\,s), measured on an NVIDIA RTX~3090 against Forewarn's more powerful RTX~A6000. Additionally on LIBERO-10, \ours{}-1B reduces average episode length by 36\% ($390 \to 250$ simulation steps, $19.8 \to 13.0$ decision steps), yielding a net episode-level overhead of only ${\sim}3$\,s over the baseline.

\section{Limitations}
\label{sec:limitations}

DILLO requires re-alignment of the projectors for each new policy architecture, as the latent manifold $z_t$ is architecture-specific. The training pipeline itself is environment- and action-representation-agnostic, enabling direct deployment on physical hardware by having the teacher annotate real trajectories. The privileged information (ground-truth 6DoF poses) is required only for the teacher during offline annotation, never at inference; in practice, it can be substituted by estimated poses using a pose estimation algorithm \cite{foundationposewen2024}, though perception noise may degrade distillation quality.

\section{Conclusion}
\label{sec:conclusion}

We introduce DILLO, a Distilled Language-Action World Model that bypasses the latency bottleneck of visual simulation in proactive agent steering. By distilling the semantic foresight of a simulator-privileged teacher into a fast, latent-conditioned student, we demonstrate that a policy's internal representations already encode the necessary dynamics to anticipate action outcomes. This Latent Sufficiency Hypothesis allows us to shift the paradigm of world modeling from computationally expensive ``simulate-then-act'' pipelines to a streamlined ``describe-then-act'' approach. 

Our extensive evaluation across both single-task (MetaWorld) and multi-task (LIBERO) environments confirms that DILLO produces high-fidelity outcome descriptions and failure verdicts directly from the policy latent $z_t$. This architecture achieves a ${\sim}14{\times}$ inference speedup over visual world models while effectively steering policies to high success rates. Ultimately, DILLO establishes that real-time semantic foresight does not require the expensive rendering of future states, offering a fast, interpretable, reliability layer for embodied agents.

\section*{Acknowledgments}
I.S. gratefully acknowledges the NVIDIA Academic Grant Program for the donation of a DGX Spark. This work was further supported by Panasonic and by Sapienza grant RG123188B3EF6A80 (CENTS). We also thank CINECA for the allocation of computational resources.

{
    \small
    \bibliographystyle{splncs04}
    \bibliography{main}
}

\clearpage
\appendix
\renewcommand{\thesection}{\Alph{section}}
\renewcommand{\theHsection}{appendix.\Alph{section}}

\section*{Supplementary Material}
\addcontentsline{toc}{section}{Supplementary Material}
We supplement the main paper with additional details, results, and qualitative examples.
To complement the Fidelity Text-to-Observation results available in the main manuscript, we present Fidelity Text-to-Text results (Sec.~\ref{supmat:f2t2}).
We also provide the input prompts used for the baselines (Sec.~\ref{supmat:prompts}), and additional qualitative examples (Sec.~\ref{supmat:qualitative}). Furthermore, we describe and show results from an AI-Based Study (Sec.~\ref{supmat:aistudy}) and a Human Validation Study (Sec.~\ref{supmat:humanstudy}). Finally, we explore alternative solutions for steering (Sec.~\ref{supmat:ablations}).

\section{Fidelity Text-to-Text}
\label{supmat:f2t2}

To complement our physical grounding reported in Sec.4, we introduce Fidelity Text-to-Text (T2T). Unlike the Text-to-Observation (T2O) metric, which compares predictions against physical state transitions, T2T evaluates the alignment between the predicted description and the ground truth reference text. We apply the parser function $g_{parse}$ to both the generated and the ground-truth descriptions, projecting both into the shared space of canonical directional labels.
The final Fidelity T2T score is computed as the accuracy ratio of matching labels for every example. We note that this is a proxy for textual faithfulness rather than physical grounding; this metric effectively quantifies how closely the model reproduces the linguistic patterns of the reference descriptions.
The Fidelity Text-to-Text results, presented in Table~\ref{tab:fidelity_t2t_merged}, demonstrate that both the \ours{}-1b and \ours{}-4b variants substantially outperform their respective baselines.

\begin{table}[H]
    \centering
    \caption{\textbf{Fidelity Text-to-Text Results.} We compare \ours{} against predictive baselines (Few-Shot and Zero-Shot) across two model scales (1B and 4B parameters) on both the MetaWorld and LIBERO suites. Unlike the predictive baselines, the captioning model (CAP) has privileged access to past and future information. \ours{} consistently outperforms the fair predictive baselines, validating that the latent state is a sufficient statistic for semantic prediction.}
    \label{tab:fidelity_t2t_merged}
    \setlength{\tabcolsep}{4.5pt}
    \resizebox{\textwidth}{!}{
    \begin{tabular}{llcccccccc}
        \toprule
        & & \multicolumn{4}{c}{\textbf{1B Parameter Models}} & \multicolumn{4}{c}{\textbf{4B Parameter Models}} \\
        \cmidrule(lr){3-6} \cmidrule(lr){7-10}
        Suite & \textbf{Task} & \textbf{\ours-1B} & FS & ZS & CAP & \textbf{\ours-4B} & FS & ZS & CAP \\
        \midrule
        \multirow{3}{*}{MetaWorld} 
        & Soccer      & \textbf{0.523} & 0.451 & 0.289 & 0.475 & \textbf{0.655} & 0.495 & 0.519 & 0.617 \\
        & Sweep-Into  & \textbf{0.693} & 0.494 & 0.202 & 0.140 & \textbf{0.649} & 0.592 & 0.567 & 0.594 \\
        & Drawer-Open & 0.531 & \textbf{0.578} & 0.245 & 0.570 & \textbf{0.735} & 0.685 & 0.245 & 0.621 \\
        \midrule
        \multirow{5}{*}{LIBERO} 
        & Goal    & \textbf{0.727} & 0.262 & 0.287 & 0.723 & 0.710 & 0.422 & 0.241 & \textbf{0.905} \\
        & Object  & 0.622 & 0.388 & 0.204 & \textbf{0.820} & 0.587 & 0.546 & 0.225 & \textbf{0.877} \\
        & Spatial & 0.668 & 0.546 & 0.341 & \textbf{0.794} & 0.656 & 0.563 & 0.194 & \textbf{0.902} \\
        & 10      & \textbf{0.648} & 0.214 & 0.199 & 0.630 & 0.686 & 0.449 & 0.265 & \textbf{0.867} \\
        & 90      & \textbf{0.708} & 0.258 & 0.206 & 0.640 & 0.650 & 0.278 & 0.258 & \textbf{0.713} \\
        \bottomrule
    \end{tabular}
    }
\end{table}

\subsection{Parser Definition}
\label{supmat:gparse}

Both T2O and T2T rely on a shared deterministic parser $g_{\text{parse}}$, which projects a free-form natural language description into a fixed set of canonical semantic slots. Given a textual description $d$, the parser $g_{\text{parse}}(d)$ returns a structured representation with two entity scopes, \emph{robot} and \emph{object}, each comprising directional and relational slots:
\begin{itemize}
    \item \textbf{Robot (arm/gripper):} three directional slots for each spatial axis (\texttt{x} $\in$ \{\textit{left}, \textit{right}, \textit{no change}\}, \texttt{y} $\in$ \{\textit{forward}, \textit{backward}, \textit{no change}\}, \texttt{z} $\in$ \{\textit{up}, \textit{down}, \textit{no change}\}) and one relational slot (\texttt{approach} $\in$ \{\textit{approaching object}, \textit{moving away from object}, \textit{no arm-to-object change}\}).
    \item \textbf{Object:} three directional slots (\texttt{x}, \texttt{y}, \texttt{z}), using the same directional labels plus \textit{stationary} for no movement, and one task-progress slot (\texttt{target} $\in$ \{\textit{on target}, \textit{closer to target}, \textit{farther from target}\}, or task-specific labels such as \textit{opening drawer}, \textit{closed drawer}).
\end{itemize}
Slots that are not mentioned in $d$ are left as \texttt{null} and excluded from scoring.

\paragraph{Parsing Procedure.}
Given a textual description $d$, the parser operates as follows:
\begin{enumerate}
    \item \textbf{Preprocessing:} All XML/HTML tags are stripped and the text is lowercased.
    \item \textbf{Sentence splitting:} The text is split into individual sentences using punctuation delimiters (\texttt{.}, \texttt{;}, \texttt{!}, \texttt{?}).
    \item \textbf{Entity-scoped extraction:} For each sentence, the parser determines whether it mentions the \emph{robot} (via aliases such as ``gripper'', ``arm'', ``end-effector'', etc.), the \emph{object} (via aliases such as ``object'', ``ball'', ``cube'', etc.), or both, using regular expression matching. Directional keywords are then attributed to the appropriate entity:
    \begin{itemize}
        \item If only the robot is mentioned, directional and approach cues are assigned to the robot slots.
        \item If only the object is mentioned, directional and target-progress cues are assigned to the object slots.
        \item If both are mentioned in the same sentence, the sentence is split on the object noun, and the first clause is attributed to the robot while the second is attributed to the object.
    \end{itemize}
    \item \textbf{Directional matching:} Within each clause, axis-aligned movement labels are extracted by matching against curated regular expression patterns. For example, the $x$-axis is labeled \textit{left} if patterns such as ``\textit{left}'', ``\textit{to the left}'', or ``\textit{$-$x}'' are matched and no contradictory rightward pattern is present. Analogous rules apply to the $y$- and $z$-axes.
    \item \textbf{Global cues:} If stationarity keywords (\eg, ``stationary'', ``motionless'', ``remains in place'') appear anywhere in the description, all object directional slots are set to \textit{stationary}.
\end{enumerate}

\begin{figure}[!htbp]
    \centering
    \begin{lstlisting}[
    basicstyle=\tiny\ttfamily, breaklines=true, frame=single
    ]
You are given several consecutive observations from a robotic manipulation episode. The robot must interact with an object to achieve a task.
Object: drawer
Task: The task is to open the drawer
Task goal: The drawer should be opened in such a way that it reaches the target position
Target position (xyz): [0., 0.539, 0.09]

Observation schema
- Gripper: [x, y, z, openness]
    * x, y, z (meters in a fixed world frame)
    * openness, with values in [0, 1] where 0=closed, 1=fully open
- Object: [x, y, z]
    * x, y, z (same frame as gripper)

Oracular schema (future hint)
- You will be given an Oracular block with auxiliary information about the immediate next observation (i.e., what happens right after the last observation).
- How to use it:
    * Treat the oracular information as privileged signal to resolve ambiguity and correct noise from the observation trend or action intent.
    * Never copy numbers or restate raw values; use only qualitative cues for a concise, natural-language description.
    * The oracle only refers to the *imminent* next step; do not speculate beyond it.

Describe what will happen next, the imminent behavior after the last given observation. Focus strictly on the robot object interaction.

What to cover (in order)
1) Proximity/intent: Is the gripper moving toward or away from the object?
2) Mode of interaction: preparing to interact, aligning, executing a grasp, positioning to push/slide/deflect, maintaining or breaking contact.
3) Object reaction: is the object starting to move (and in which direction)? If the robot is not in contact, note residual motion (e.g., momentum transfer).
4) Gripper openness change: closing to grasp intent; opening to release/avoid; steady to maintain/wait.

Direction words
- Left/Right: x decrease/increase
- Backward/Forward: y decrease/increase
- Down/Up: z decrease/increase
Never say "along the X-axis/Y-axis/Z-axis"; use the words above.

Style constraints
- First describe the robotic arm, then the object.
- Keep it 1-3 sentences, concise, action-focused.
- Avoid hedging (e.g., "maybe", "might") unless evidence is weak.
- If the arm moves away and the object is not moving (or moving away from the target), point out potential misbehavior.
- Do not include numbers, calculations, or restate the raw observations.
- Output only the description text.

Example 1:
- Obs 1 Gripper: [-0.158, 0.851, 0.402, 0.851] Object: [-0., 0.728, 0.09]
- Oracular Info Gripper: [-0.138, 0.852, 0.415, 0.718] Object: [-0., 0.728, 0.09]
- Description: The gripper moves slightly right, forward, and up while closing, suggesting it's attempting to align with and grasp the object. The object remains stationary.

Now produce the description for the following input:
- Obs 1 Gripper: [0.001, 0.736, 0.154, 1.] Object: [0., 0.701, 0.09]
- Oracular Info Gripper: [-0.006, 0.696, 0.146, 0.979] Object: [0., 0.662, 0.09]

The description is:
\end{lstlisting}
    \caption{Gemma-3-1b and Gemma-3-4b prompts used in the Fewshot-Captioning configuration. For Gemma-3-4b, we also interleave images with the given observations.}
    \label{fig:oracular_prompt}
\end{figure}

\section{Baselines Prompts}
\label{supmat:prompts}
All baselines utilize a shared base prompt, identical to the one employed for the ZeroShot baseline, in which we explicitly define the input schema and instruct the model to generate a description. This schema ensures the model correctly interprets the numerical observation data by specifying the following task-dependent elements:
\begin{itemize}
    \item \textbf{Object Type:} The entity involved in the interaction.
    \item \textbf{Task Description:} A high-level definition of the objective, such as minimizing the distance of an object to a target position, opening a drawer, or moving a ball into a goal.
    \item \textbf{Task Goal:} The specific success condition (\eg, ``The drawer should be opened to reach the target position'').
    \item \textbf{Target Position:} The coordinates required to enable reasoning about object-to-target relationships.
\end{itemize}

To adapt this for the FewShot baseline, we extend the prompt by appending three input-output demonstrations.
Finally, for the FewShot-Captioning baseline, we supplement both demonstrations and inputs with information of the future. We introduce an explicit schema to instruct the model on how to utilize this privileged data, defined as the observation of the environment state following the execution of an action.
Figure \ref{fig:oracular_prompt} illustrates the complete prompt used for the Fewshot-Captioning baseline; note that only a single demonstration is displayed due to space constraints. Additionally, for the 4b baselines, we modify the input prompt by interleaving RGB images with their corresponding observations.

\section{Additional Qualitative Examples}
\label{supmat:qualitative}
\begin{figure}[H]
  \centering
  \includegraphics[width=\textwidth]{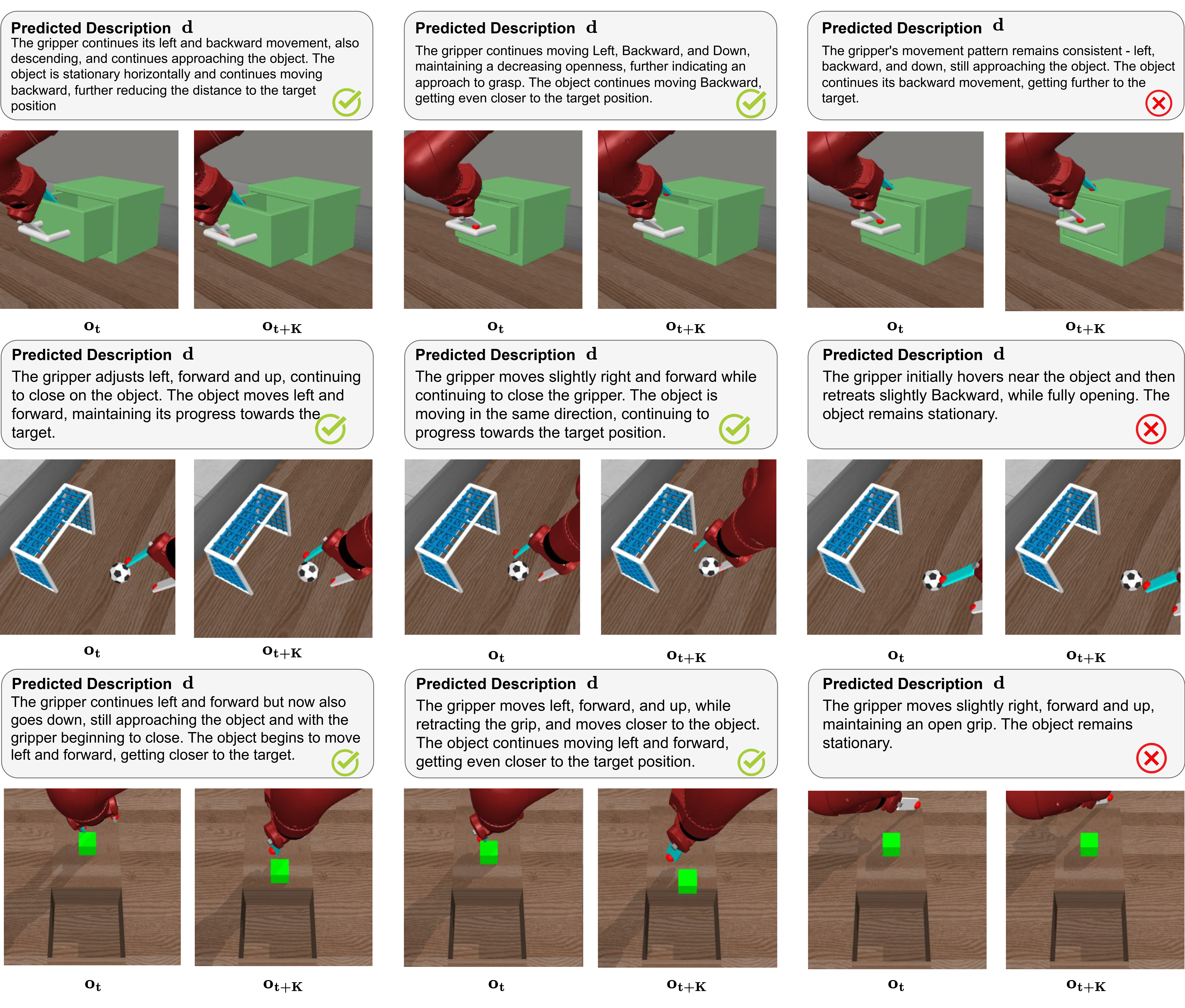}
  \caption{\textbf{Qualitative examples of DILLO for Drawer-Open (Top), Soccer (Middle), and Sweep-Into (Bottom) tasks.}}
  \label{fig:add_qual}
\end{figure}

We provide comprehensive qualitative examples covering the three tasks utilized from the MetaWorld simulator in Figure~\ref{fig:add_qual}.
Note that the gripper and object directional changes in the descriptions must be interpreted from the robotic arm's point of view.
Our model, \ours{}, successfully describes the motion of both the gripper and the object of interest, while also accurately accounting for crucial gripper-object and object-target relationships (\eg, noting that the object is ``further from the target'' in the negative example for the Drawer-Open task). \ours{}, correctly predicts the verdict (indicated by green and red symbols in the figure). A verdict of \textsc{Negative} is specifically triggered when the robotic arm moves away from the target and object positions.

\begin{figure}[H]
  \centering
  \includegraphics[width=\textwidth]{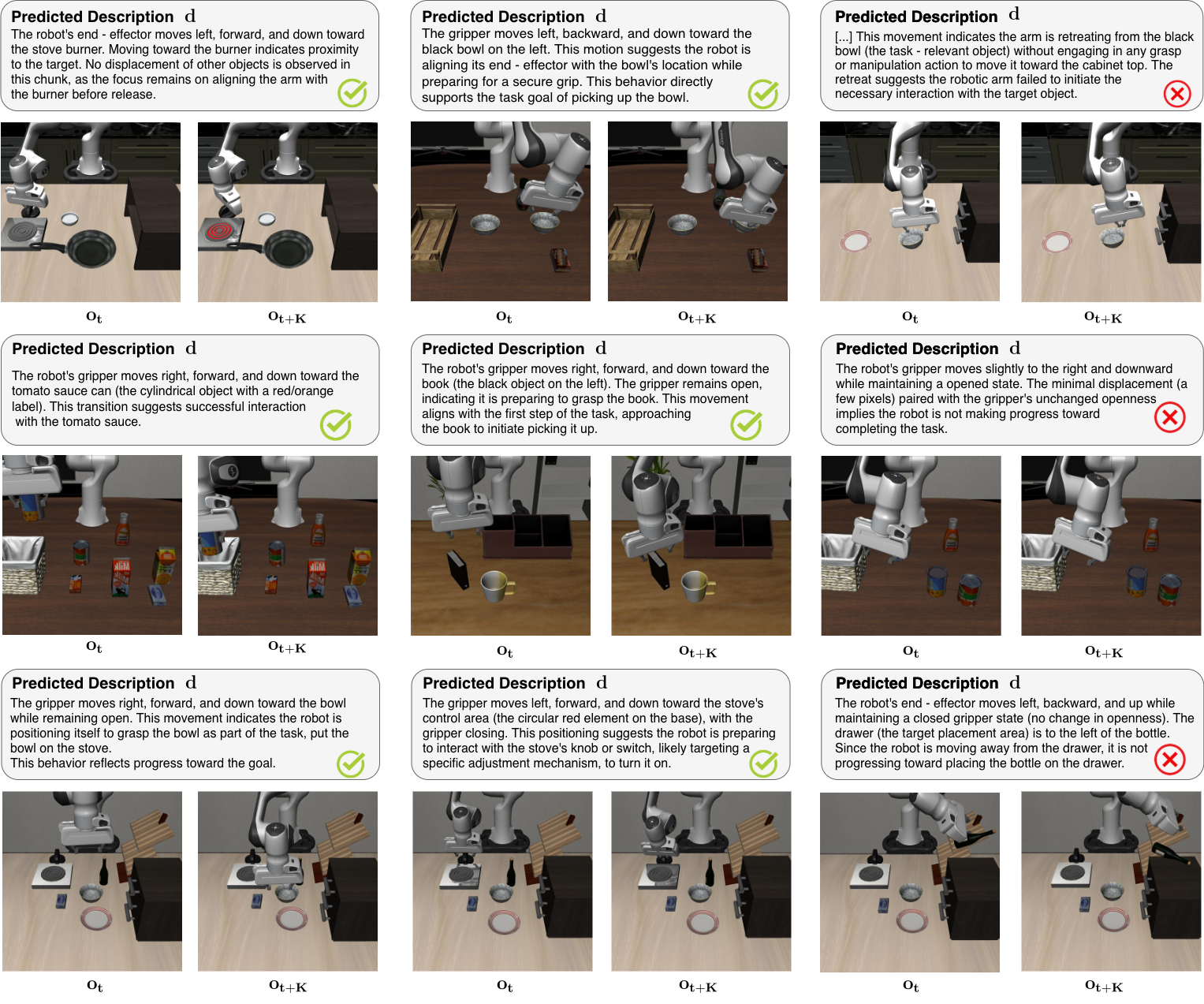}
  \caption{\textbf{Qualitative examples of DILLO for LIBERO-90 (Top), 
  LIBERO-10 (Middle), and LIBERO-Goal (Bottom) tasks. Descriptions are trimmed 
  for readability.}}
  \label{fig:add_qual_libero}
\end{figure}

In Figure~\ref{fig:add_qual_libero}, we report additional qualitative rollouts on LIBERO-90, LIBERO-10, and LIBERO-Goal. Compared to MetaWorld, these tasks involve more objects, longer horizons, and language-conditioned goals; as a result, the LIBERO-trained \ours{} naturally produces more detailed and fine-grained descriptions that capture multi-step intent and object–object relations. For readability, we truncate these descriptions in the figure—for instance, the middle \textsc{Positive} example in LIBERO-Goal would read:\\

\textit{"The gripper moves left, forward, and down toward the stove's control area (the circular red element on the base), with the gripper closing. This positioning suggests the robot is preparing to interact with the stove's knob or switch, likely targeting a specific adjustment mechanism, to turn it on. The small movement in the gripper's path of least resistance aligns with approaching the task-relevant object (stove control), indicating intent to engage with the system for activating the stove. No visible displacement of the stove itself occurs, but the directional shift indicates preparation for direct interaction with the ignition component."},\\

which is shortened to a concise summary in the visualization.

\section{AI-Based Study}
\label{supmat:aistudy}
To evaluate the quality and coherence of the generated descriptions across our model and baselines, we conducted an AI-based preference study using an LLM as an automated judge.
This approach allows us to measure the perceived accuracy of the linguistic outputs at scale, mitigating the expense and variability of traditional human studies.
The study relies on a specific, structured prompt designed to instruct the judging LLM (Qwen2.5-32B-Instruct) on the preference task. Each query includes the following inputs:
\begin{itemize}
    \item Observations $(o_t, o_{t+k})$: Two consecutive observations (current state and subsequent state) representing the physical transition.
    \item Candidate Descriptions: A set of $K+1$ descriptions, consisting of one Ground Truth (GT) description and $K$ descriptions generated by the baselines and \ours{}. Each candidate is assigned a unique, numeric ID (from 1 to $K$).
    \item Instructions: Guidance on the reference example and constraints for the output format (returning only the preferred ID).
\end{itemize}
The employed query prompt is shown in Figure~\ref{fig:ai_study_prompt}.
To automate the evaluation process, we define a function responsible for prompt formatting, querying the LLM judge, and extracting the preference ID. To mitigate positional bias (\ie, preventing the model from systematically selecting the first ID), the candidate IDs and their corresponding descriptions are shuffled prior to formatting. 
The results of this AI-based study are detailed in Figure~\ref{fig:ai_study}. As can be seen, both the \ours{}-1b and \ours{}-4b models achieve a higher judge preference compared to the baselines.
\clearpage
\begin{figure}[H]
    \centering
    \begin{lstlisting}[
    basicstyle=\scriptsize\ttfamily, breaklines=true, frame=single
    ]
You are a meticulous evaluator. You are given a REFERENCE (ground-truth description)
and several MODEL RESPONSES describing the same event. Compare them carefully and decide
which numbered response best matches the REFERENCE at the level of atomic actions
(motion, direction, gripper state, object behavior, etc.).

## Rules
- Focus on how accurately each response reflects the REFERENCE.
- Ignore harmless extra details.
- Penalize contradictions (e.g., left vs right, open vs close).
- Choose exactly ONE winner (the best match).
- If two are equally good, choose the more precise one.

## Output Format
Return ONLY valid JSON in this exact format:

{{
  "example_id": "<EXAMPLE_ID>",
  "winner_index": <int>,  // index of the best response (1-based)
  "reason": "<short justification>"
}}

Now evaluate this case.

EXAMPLE_ID: {example_id}

REFERENCE:
{reference}

RESPONSES:
{responses}
\end{lstlisting}
    \caption{The input prompt utilized for querying Qwen2.5-32B-Instruct in our AI-based study}
    \label{fig:ai_study_prompt}
\end{figure}

\begin{figure}[htbp]
    \centering
    \includegraphics[width=0.7\linewidth]{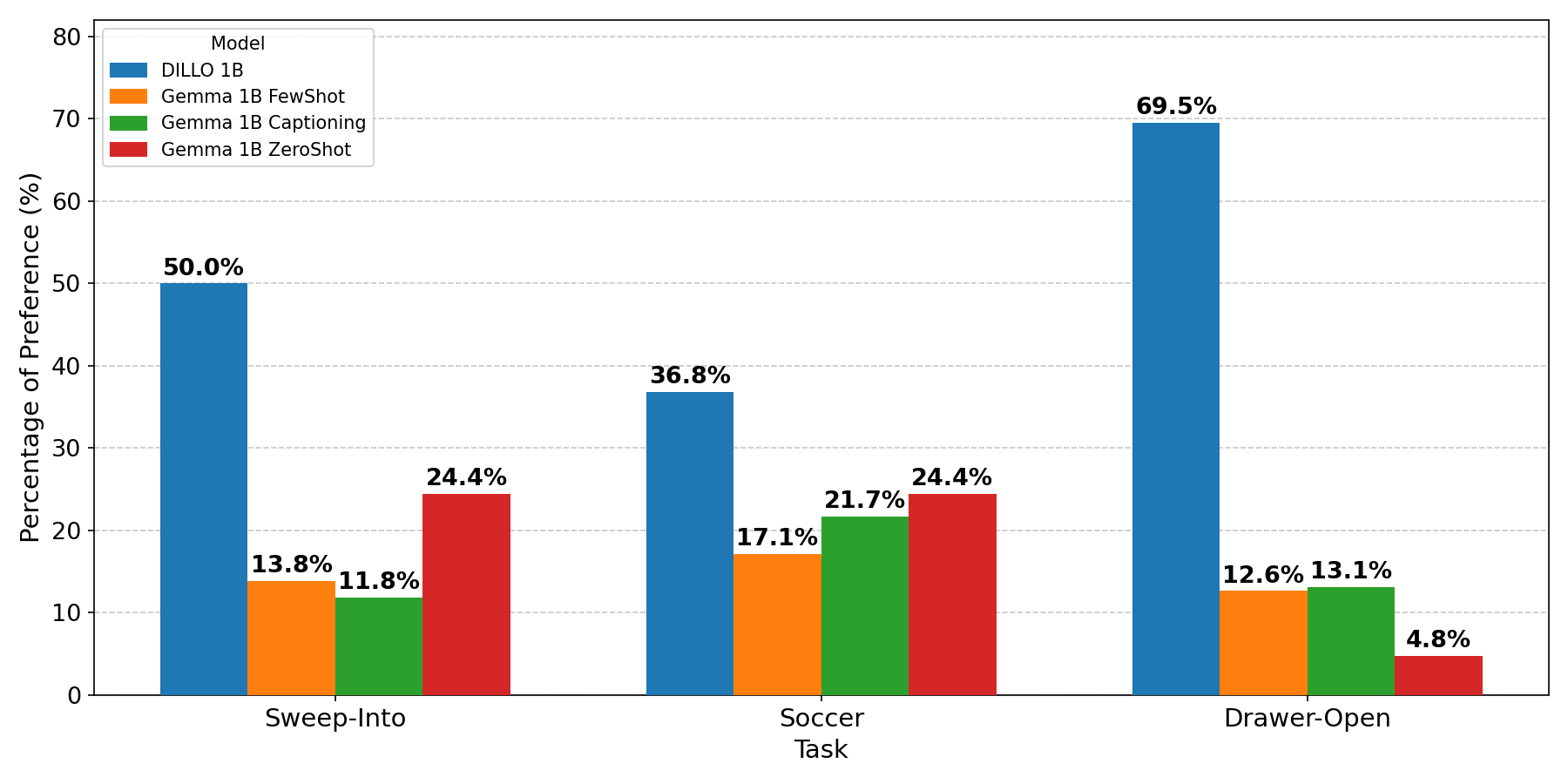} \\
    \includegraphics[width=0.7\linewidth]{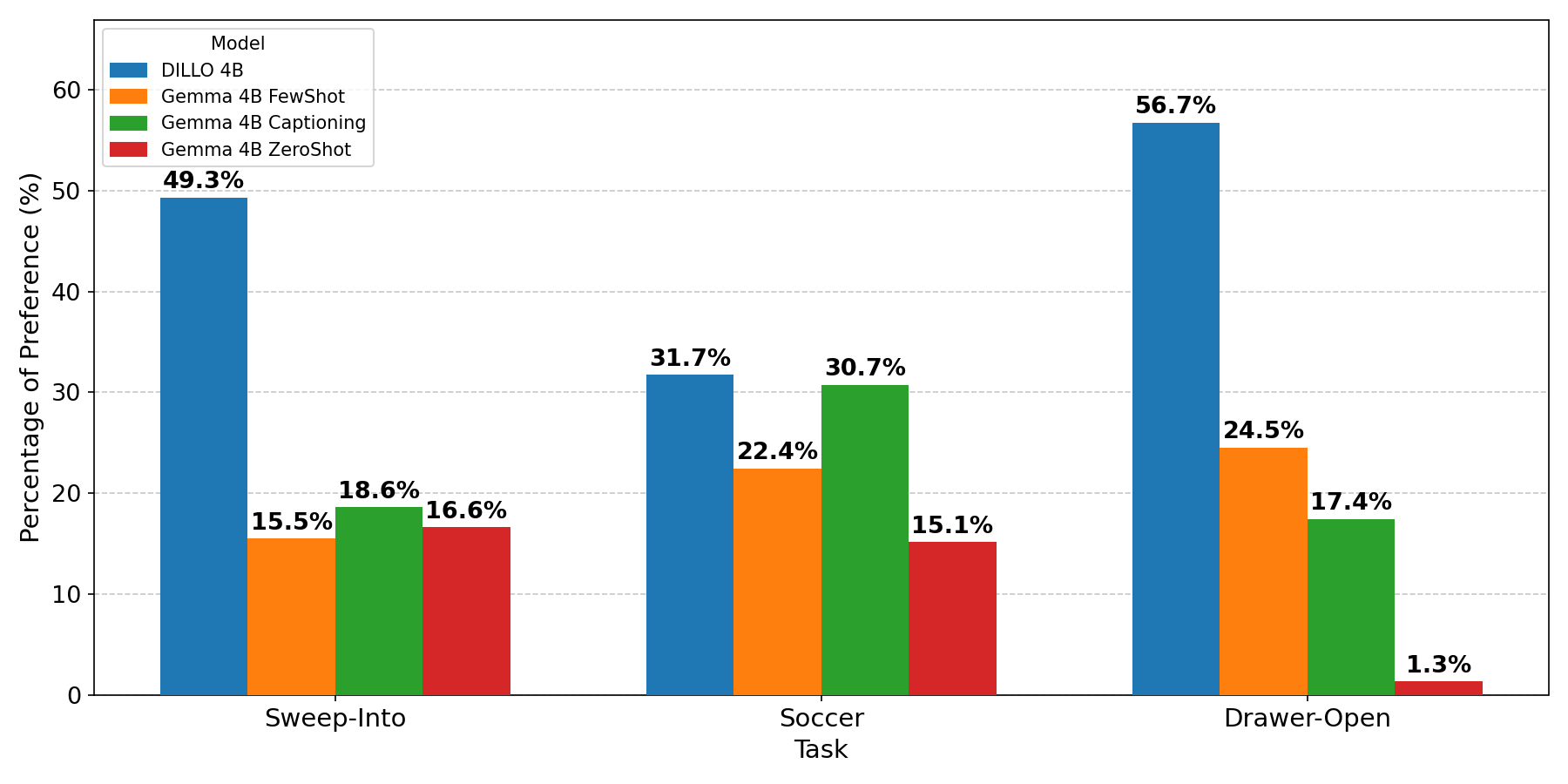} \\
    \includegraphics[width=0.7\linewidth]{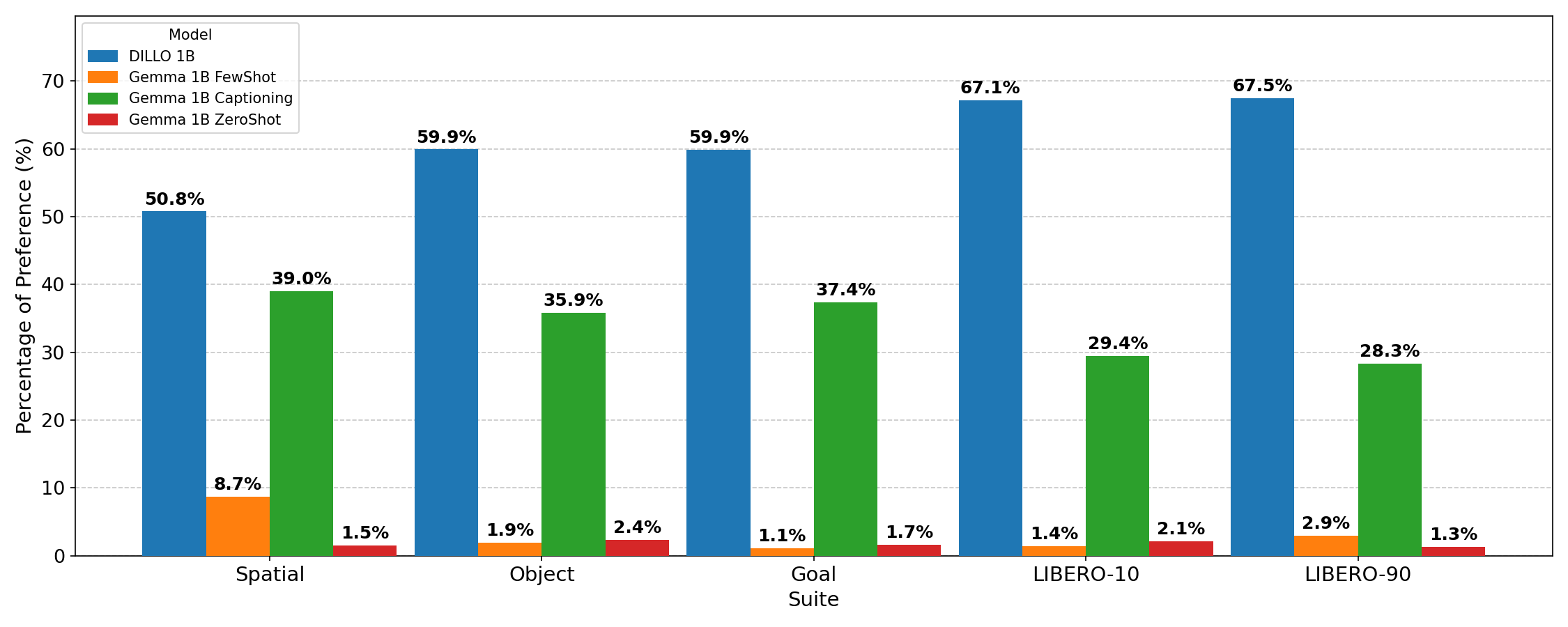} \\
    \includegraphics[width=0.7\linewidth]{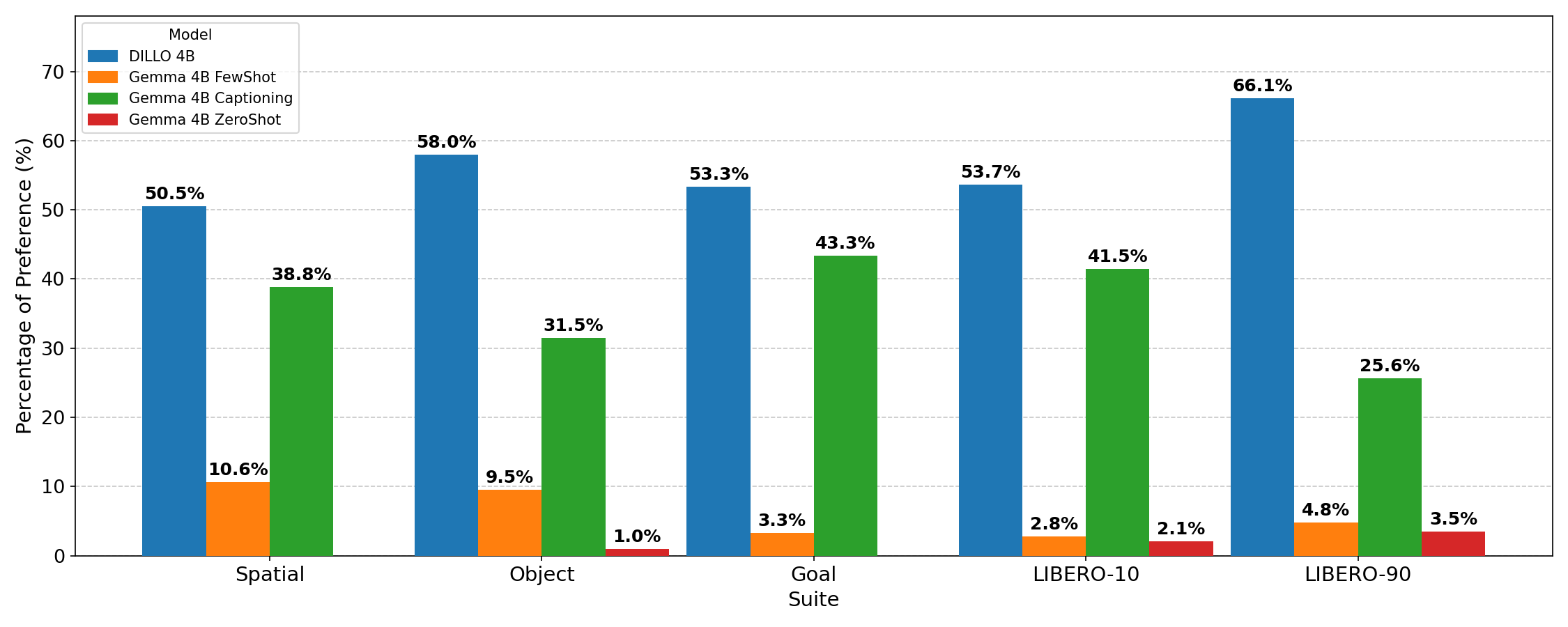}
    
    \caption{AI judge preference percentage results for the 1B and 4B models across the three MetaWorld tasks (Sweep-Into, Soccer, and Drawer-Open) and the five LIBERO suites (Spatial, Object, Goal, LIBERO-10, and LIBERO-90).}
    \label{fig:ai_study}
\end{figure}

\clearpage

\section{Human Validation Study}
\label{supmat:humanstudy}
To complement the automated evaluation, we conducted a human validation study on 120 randomly sampled episodes: 60 from MetaWorld (20 per task, balanced between \textsc{Positive} and \textsc{Negative} outcomes) and 60 from LIBERO.
\noindent\textbf{Annotators.}
Ten annotators with a background in robotics and computer vision (PhD-level) participated in the study. Prior to participation, all annotators were provided with an informed consent form detailing the study's purpose, task structure, and data usage policy, and confirmed their agreement before proceeding.
\noindent\textbf{Protocol.}
For each episode, annotators were shown: (i) the natural language task instruction, (ii) \ours{}'s generated description $d_{t+k}$, and (iii) the RGB observations before and after action execution ($o_t$ and $o_{t+k}$). Annotators were then asked to:
\begin{enumerate}
    \item Rate the accuracy of $d_{t+k}$ on a 1--5 Likert scale, where 
    1 = \emph{Completely Inaccurate} and 5 = \emph{Perfectly Accurate}.
    \item Classify the observed outcome as \textsc{Positive} (effective, 
    advances the task) or \textsc{Negative} (ineffective, useless, or 
    detrimental to the goal).
\end{enumerate}
Human verdict labels were then compared against \ours{}'s predicted verdict $\hat{c}$ and the ground-truth labels $c_T$. 
Formally, given per-episode labels $\hat{c}_i$ from \ours{} and $c^{\text{human}}_i$ from the aggregated human majority vote, we define agreement as
\[
\text{Agree}(\hat{c}, c^{\text{human}}) \;=\; \frac{1}{N} \sum_{i=1}^{N} 
\mathbf{1}\big[\hat{c}_i = c^{\text{human}}_i\big],
\]
with analogous definitions for \(\text{Agree}(c_T, c^{\text{human}})\) and \(\text{Agree}(\hat{c}, c_T)\).
\noindent\textbf{Results.}
On MetaWorld, \ours{}'s predicted verdict $\hat{c}$ agrees with human assessments on \textbf{84.1\%} of episodes. On LIBERO, \ours{} achieves a verdict agreement with ground truth $c_T$ of \textbf{72.1\%}, exceeding the human-vs-ground-truth agreement of 69.5\%, indicating that \ours{}'s latent-conditioned predictions are a more reliable safety signal than human visual inspection alone. \ours{}-vs-human agreement on LIBERO stands at 68.8\%. Regarding description quality, human annotators assigned an average accuracy rating of $\mathbf{3.62/5.0} \pm 1.09$, confirming that the distilled natural language foresight is both accurate and interpretable to domain-expert observers.

\section{Ablation Studies}
\label{supmat:ablations}

We investigate which components of \ours{} contribute to effective policy steering by comparing the complete multistage training procedure against three alternatives. \emph{Stage 3 only} directly optimizes the final description-and-verdict objective, omitting the projector-alignment and description-distillation stages. \emph{Verdict only} removes semantic description generation and trains the model to predict only the binary \textsc{Positive}/\textsc{Negative} verdict. Finally, the \emph{MLP critic} directly predicts the binary verdict from the policy latent and candidate action chunk, providing a simpler learned classifier without language generation. All variants are evaluated using the same rejection-sampling protocol and candidate budget (N=5).

\begin{table}[t]
\centering
\caption{\textbf{Steering ablation on LIBERO validation suites.}
Episode success rates (\%). Multistage denotes the complete DILLO training procedure.}
\label{tab:steering_full_ablation}
\small
\setlength{\tabcolsep}{3.5pt}
\renewcommand{\arraystretch}{1.08}

\resizebox{\linewidth}{!}{\begin{tabular}{l c ccc ccc c}
    \toprule
    & & \multicolumn{3}{c}{DILLO-1B} & \multicolumn{3}{c}{DILLO-4B} & \\
    \cmidrule(lr){3-5}\cmidrule(lr){6-8}
    Suite & Base & Multistage & \shortstack{Stage 3\\only} & \shortstack{Verdict\\only}
    & Multistage & \shortstack{Stage 3\\only} & \shortstack{Verdict\\only} & MLP \\
    \midrule
    LIBERO-Goal    & 86.0 & 87.5 & 79.0 & 85.0 & 89.0 & \textbf{92.0} & 83.0 & 86.0 \\
    LIBERO-Object  & 82.0 & 86.5 & 80.0 & 83.0 & \textbf{88.0} & 81.0 & 81.0 & 80.5 \\
    LIBERO-Spatial & 70.0 & \textbf{76.0} & 69.0 & 71.0 & 73.0 & 54.0 & 64.0 & 70.5 \\
    LIBERO-10      & 64.0 & 71.5 & 59.0 & 65.0 & \textbf{74.0} & 67.0 & 63.0 & 64.0 \\
    LIBERO-90      & 71.4 & 68.6 & 70.8 & 71.5 & 70.5 & \textbf{73.6} & 70.1 & 70.2 \\
    \bottomrule
\end{tabular}}

\end{table}

Table~\ref{tab:steering_full_ablation} shows that the complete multistage procedure achieves the strongest overall steering performance for both model sizes. Averaged across the five LIBERO suites, \ours{}-1B and \ours{}-4B obtain success rates of (78.0\%) and (78.9\%), respectively, compared with (74.7\%) for the base policy. Removing the training curriculum substantially reduces performance: the Stage-3-only variants achieve (71.6\%) and (73.5\%), corresponding to decreases of (6.4) and (5.4) percentage points relative to the complete 1B and 4B models. Although Stage-3-only training performs strongly on individual suites, such as LIBERO-Goal and LIBERO-90 for the 4B model, its performance is less consistent across tasks. This indicates that directly optimizing the final objective can produce effective task-specific steering, but does not provide the same robustness as the progressive alignment and distillation procedure.

The verdict-only ablation further isolates the contribution of semantic description generation. Without the description objective, average success decreases from (78.0\%) to (75.1\%) for the 1B model and from (78.9\%) to (72.2\%) for the 4B model. The effect is particularly pronounced for \ours{}-4B, suggesting that binary supervision alone does not reliably exploit the capacity of the larger language model. Generating a description of the predicted physical outcome provides a denser intermediate supervision signal, encouraging the model to ground its verdict in the expected consequences of the proposed action rather than learning only a binary decision boundary.

The latent-conditioned MLP critic also fails to improve the base policy on average, obtaining (74.2\%) success compared with the base policy's (74.7\%). It leaves performance unchanged on LIBERO-10 and LIBERO-Goal, produces only a marginal improvement on LIBERO-Spatial, and decreases performance on LIBERO-Object and LIBERO-90. Thus, access to the policy latent, candidate actions, and binary teacher labels is not sufficient to produce reliable steering with the tested classifier. Overall, these ablations show that both the multistage curriculum and semantic outcome-generation objective contribute to \ours{}'s steering performance. While they do not establish that natural-language generation is strictly necessary, they indicate that semantic description generation provides a useful intermediate learning signal that leads to more consistent and effective action selection than direct binary classification.

\end{document}